\documentclass[conference]{IEEEtran}
\IEEEoverridecommandlockouts
\usepackage{cite}
\usepackage{amsmath,amssymb,amsfonts}
\usepackage{algorithmic}
\usepackage{graphicx}
\usepackage{textcomp}
\usepackage{xcolor}
\usepackage{multirow}
\usepackage{enumitem}   
\usepackage{url}
\usepackage{hyperref}
\usepackage{comment}
\usepackage{float}
\begin{document}

\title{\title{No Rumours Please! A Multi-Indic-Lingual Approach for COVID Fake-Tweet Detection}*\\
{\footnotesize \textsuperscript{*}Note: Sub-titles are not captured in Xplore and
should not be used}
\thanks{Identify applicable funding agency here. If none, delete this.}
}

 \author{\IEEEauthorblockN{Debanjana Kar*}
 \IEEEauthorblockA{\textit{Dept. of CSE} \\
 \textit{IIT Kharagpur, India}\\
 debanjana.kar@iitkgp.ac.in}
 \and
 \IEEEauthorblockN{Mohit Bhardwaj* \thanks{* Work done during internship at IBM Research, India.}}
 \IEEEauthorblockA{\textit{Dept. of CSE} \\
 \textit{IIIT Delhi, India}\\
mohit19014@iiitd.ac.in}
 \and
 \IEEEauthorblockN{Suranjana Samanta}
 \IEEEauthorblockA{\textit{IBM Research} \\
 \textit{Bangalore, India}\\
 suransam@in.ibm.com}
 \and
 \IEEEauthorblockN{Amar Prakash Azad}
 \IEEEauthorblockA{\textit{IBM Research} \\
 \textit{Bangalore, India}\\
 amarazad@in.ibm.com}

}

\maketitle

\begin{abstract}
% Fake news/tweets, specially about a sensitive matter, can bring negative rippling effect in our society. This work describes a method to detect fake tweets about COVID19 pandemic in a multilingual environment. We use a rich feature set, consisting of text embedding along with relevant hand-crafted features capturing the characteristics of the Tweet text and the Twitter user, to identify fake tweets. The method is language agnostic and we have experimented with tweets in English, Hindi and Bengali. We create a new tweet dataset by expanding the Covid19 Twitter dataset \cite{infodemic} with our scrapped Hindi and Bengali tweets regarding Covid19. Experiments on the dataset shows the effectiveness of our proposed approach in detecting a fake tweet.
The sudden widespread menace created by the present global pandemic COVID-19 has had an unprecedented effect on our lives. Man-kind is going through humongous fear and dependence on social media like never before. Fear inevitably leads to panic, speculations, and spread of misinformation. Many governments have taken measures to curb the spread of such misinformation for public well being. Besides global measures, to have effective outreach, systems for demographically local languages have an important role to play in this effort. Towards this,  we propose an approach to detect fake news about COVID-19 early on from social media, such as tweets, for multiple Indic-Languages besides English.  In addition, we also create an annotated dataset of Hindi and Bengali tweet for fake news detection. % (being released). 
We propose a BERT based model augmented with additional relevant features extracted from Twitter to identify fake tweets.  To expand our approach to multiple Indic languages, we resort to mBERT based model which is fine tuned over created dataset in Hindi and Bengali. We also propose a zero shot learning approach to alleviate the data scarcity issue for such low resource languages.  Through rigorous experiments, we show that our approach reaches around 89\% F-Score in fake tweet detection which supercedes the state-of-the-art (SOTA) results. Moreover, we establish the first benchmark for two Indic-Languages, Hindi and Bengali. Using our annotated data, our model achieves about 79\% F-Score in Hindi and 81\% F-Score for Bengali Tweets. 
Our zero shot model achieves about 81\% F-Score in Hindi and 78\% F-Score for Bengali Tweets without any annotated data, which clearly indicates the efficacy of our approach. 
\end{abstract}


\section{Introduction}
% 1 page including literature
% motivation - why fake news mitigation is important - Detection plays role
% challange in detection
% today's world cosmo society are multilingual and mixed -> multilingual model
% Challanges in multilingual Detection 

With the insurgence of the most devastating pandemic of the century, COVID-19, the entire planet is going through an unprecedented set of challenges and fear. Every now and then new revelations surface either for COVID solutions e.g. medicines, vaccines, mask usage, or regarding COVID  dangers. Along with factual information, it has been observed that large amounts of misinformation are circulating on social media platforms such as Twitter. %Spread of such misinformation or rumors brings another set of challenges to governance besides the core COVID firefight. 
The COVID-19 outbreak has affected our lives in a significant way. Not only does it pose a threat to the physical health of an individual, but rumors and fake facts can have an adverse effect on one's mental well-being. 
Such misinformation can bring another set of challenges to governance if not detected in time especially due to it's viral nature.%lightening spread capabilities of social network, e.g. Facebook and Twitter. %All such networks are actively monitoring, especially Twitter, for such information and taking preventive  actions. 
%Recently, Twitter account of US president has been blocked due to alleged misinformation (\href{https://www.bbc.com/news/election-us-2020-53673797}{ BBC News}). 
\begin{figure}[htb]
\centering
\includegraphics[scale=0.4]{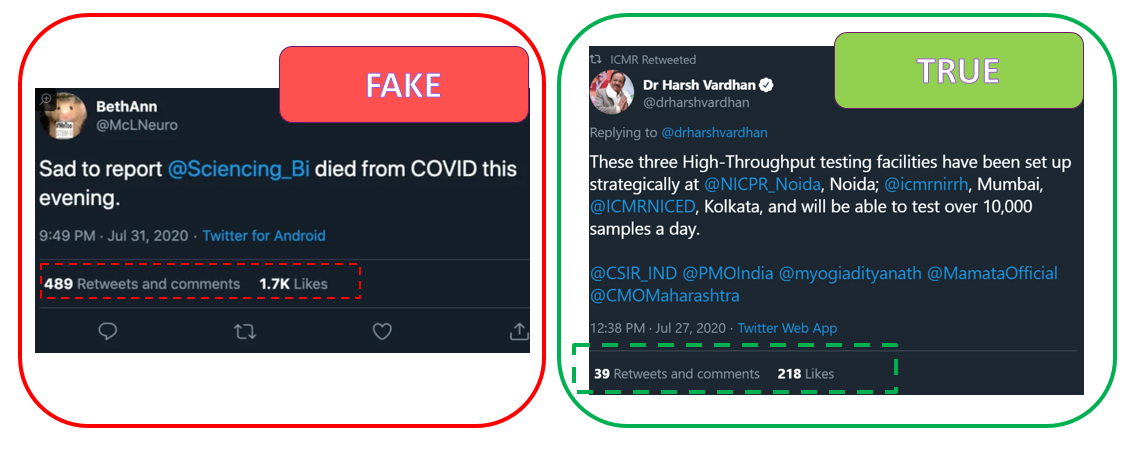}
\centering
\vspace{-0.1cm}
\caption{Two examples of COVID-19 related tweets; Left one shows an example of fake tweet, which has high number of retweet and like counts; Right one shows an example of tweet with real facts, which is has low number of retweets. Number of retweets and likes is proportional to the popularity of a tweet.}
\label{fig:example}
\vspace{-0.25cm}
\end{figure}

Veracity of rumours over social media makes the detection challenging, and has been studied widely in recent past \cite{Shu17,Ding20}. %\cite{Shu17,Yang20,Ding20}. 
Though more popular or highly retweeted messages may seem to be factual, %but 
it need not be the case especially in fast dissemination periods such as the COVID era; Fig. \ref{fig:example} depicts an example. %time  %. Observe in Fig.\ref{fig:example}, a fake message has got immense popularity as compared to another genuine message.
In addition, the proliferation of Twitter to diverse demography  induces %another 
 challenges due to usage of locality specific languages. For example, various Indic languages, e.g. Hindi, Bengali, Telugu, Kannada are widely used in Twitter. Though there are some datasets released in English \cite{infodemic} for COVID fake-news, there are hardly any datasets released for Indic-languages.

In this paper, we propose an approach where besides network and user related features, the text content of message is given high importance. In particular, we train our model to understand the textual content using mBERT \cite{devlin2018bert} on COVID dataset to classify fake or genuine tweet. 
Our proposed work describes a method to detect fake and offensive tweets about COVID-19 using a rich feature set and a  deep neural network classifier. 
We leverage Twitter dataset for fake news detection in English language released in \cite{infodemic}. Since today's society is cosmopolitan and multi-lingual, we consider a language agnostic model to detect fake tweets. 
For fake detection in a multi-lingual environment we fine-tune mBERT (Multilingual BERT)%\cite{mBertgit}
\footnote{\url{https://github.com/google-research/bert/blob/master/multilingual.md}} to obtain textual features from tweets. 
We created fake news dataset in two Indic-Languages (Hindi and Bengali) by extracting tweets and annotated them for misinformation. Our multilingual model achieves reasonably high accuracy to detect fake news in Indic-Languages when trained with combined dataset (English and Indic-Languages). %Moreover, it also improved accuracy in English language which can be attributed to rich and diverse information injection through syntactic constructs from other languages.  
Towards scalability and generalization to other Indic-Languages, we also propose a zero-shot approach where the model is trained on two languages, e.g. English with Hindi, and tested on a third language, e.g. Bengali. We experimented rigorously with various languages and dataset settings to understand the model performance.  Our experimental results indicate comparable accuracy in zero shot setting as well. Belonging to the Indo-Aryan family of languages, the Indic languages share similar syntactic constructs which seems to aid the cross lingual transfer learning and help attain good accuracy.
%As the Indic-Languages, especially Aryan family, has similar syntactic construct, the cross lingual transfer learning seems to help for such accuracy. Given the urgent need of time, we have also released an API and source code that can be integrated with an application for fake news detection. 

% We live in a world where distance seems to be a decreasing factor to be able to connect to the human population. With the ever increasing popularity of various social media platforms, everyone has the power to reach out and influence a large population in our society. A false news can have a significant negative effect on the society, which can confuse people. This can lead to a serious situation, specially in the time of pandemic. The COVID19 outbreak has affected our lives in a significant way. Not only does it pose a threat to the physical health of an individual, but rumours and fake facts can have an adverse effect on one's mental well-being. Since today's society is cosmopolitan and multi-lingual, we consider a language agnostic model to detect fake tweets. The proposed work describes a method to detect fake and offensive tweets about COVID19 using a rich feature set and a stacking ensemble deep neural network classifier.

% In our work, we emphasise on the tweet text and user information to detect the fake tweets. This is mainly due to the nature of the Covid-19 tweets that are available to train any classification model. Our aim is to handle multi-lingual tweets in a robust way which is essential for its practical usability in a multi-lingual, multi-cultural country.

Our \textbf{key contributions} can be enumerated  as follows: \\
i) We have created COVID-19 fake multilingual tweet dataset - \textit{Indic-covidemic fake tweet dataset}, for Indic Languages (Hindi and Bengali), which is being released%\footnote{\url{https://github.com/DebanjanaKar/Covid19\_FakeNews\_Detection}}
. As per our knowledge, this is the first multilingual COVID-19 tweet related dataset in Indic language. \\
ii) We propose an mBERT based model for Indic languages, namely Hindi and Bengali, for fake tweet detection. We show that the model, fine-tuned on our proposed multilingual dataset, outperforms single language models. Moreover, we establish bench mark results for Indic languages for COVID fake tweet detection. \\
iii) Our zero shot setting, suitable for low resource Indic languages, performs comparable to models trained with combined dataset on Indic languages. Our experimental evaluation indicates that feature representations are highly transferable across Indic Languages enabling extension to other Indic languages.

\section{Related literature}
\label{s:rel}
% %Literature
% Fake news detection in social media is a popular topic of interest in the field of Natural Language Processing (NLP). Since tweets are short texts with profound use of short-forms and often improper grammar, it needs to be handled in a different way than that of news articles. Infodemic Covid19 dataset \cite{infodemic} is one of the first tweet dataset to distinguish fake and negative tweets. Various transformer based models have been used for the classification task, and the comparative study shows that multilingual BERT (m-BERT) is giving the best result in classifying the tweets. Graph-aware Co-Attention Networks - GCAN \cite{lu2020gcan} is another recent work on detecting fake tweets. It uses the sequence of retweets to detect the fake tweets using graph networks. SpotFake \cite{spotfake} uses multimodal information from tweet text, using BERT embeddings, and correcponding image to detect fake tweets. Other notable work in detecting fake tweets are HawksEye \cite{hawkeseye} - which uses textual and temporal information about retweets, \cite{shu2018fakenewsnet}, \cite{hanselowski-etal-2018-retrospective} etc.

{\bf Fake News Detection:}
Fake news or false information involves various research streams such as fact-checking\cite{Thorne18}, topic credibility \cite{Thorne19}. An overview of fake news detection approaches through data mining perspective on social media has been discussed in\cite{Shu17} and \cite{Li16}. Various studies has also been carried out related to COVID-19 related disinformation.  Identifying low-credibility information using data from social media is studied in \cite{Yang20}, detecting prejudice in \cite{Vidgen20}, related to ethical issues in \cite{Ding20}, misinformation spread of socio-cultural issues in \cite{leng20} and detecting misleading information and credibility of users who spread it in \cite{Mourad20}. 
Some recent work have focussed on detecting fake news from Tweets. In GCAN \cite{lu2020gcan}, credibility of user is studied based on sequence of retweets using graph networks. On the other hand, SpotFake \cite{spotfake} uses multimodal information from tweet text, using BERT embeddings, and corresponding image to detect fake tweets. In HawksEye \cite{hawkeseye},  textual and temporal information about retweets are leveraged. 
Our work differs from these as we train our model to identify fake news leveraging mainly text features supported by other user informations related to their credibility on Twitter. 

BERT  \cite{devlin2018bert} has become a state-of-the-art as a contextual representation of various NLP tasks. mBERT, a multilingual variant, is pretrained on 104 languages which gained large popularity due to its exceedingly well performance in various cross lingual NLP tasks\cite{Pires19} lately. In our models, we finetuned both BERT and mBERT  on exiting and our created dataset for fakenews detection tasks. 
%Bert and mB based pertained models \cite{BERT} to extract sentence embeddings for various classification task has become state-o-the-art, which is often fine-tuned for specific applications. mBert \cite{BERT}, a multi-lingual BERT, pretrained on 104 lanaguges can be applied to cross-lingual environment\cite{Pires19}. 
%% zeroshot

 {\bf Covid-19 Datasets:}
Recent surge in fake news detection research is also evident from various rumour detection datasets, such as,  SemEval2019 task7\cite{Gorrel19} to determine rumour veracity, SemEval 2019 task 8 \cite{Mihaylova19} for fact-checking. Infodemic Covid-19 dataset \cite{infodemic} is one of the first tweet dataset to distinguish fake and negative tweets.
Though most studies and dataset focusses only English language, it has become more important to study in other languages as well, in multilingual, especially with regard to COVID. In \cite{Zarei20}, COVID-19 Instagram dataset is developed for multilingual usage. In \cite{infodemic}. besides English tweets, Arabic tweets have been also used to explore misinformation impact.
 %In addition \cite{coca1} is also covid related dataset for. 
However, there is no dataset available for fake tweet detection in Indic languages to the best of our knowledge. We are the first to create dataset for fake news in Twitter in two common Indic Languages, Hindi and Bengali, called as Indic-covidemic. %In this paper, we use our annotated dataset. 

\section{Data set}
\textit{Indic-covidemic tweet dataset} is one of the first multilingual Indic language tweet dataset designed for the task of detecting fake tweets. We describe the details of the dataset in the following sub-sections.

\subsubsection{English Annotations}
In our work, the definition of fake tweets holds as follows : \textit{"Any tweet that doesn't contain a verifiable claim is a malicious or fake tweet."} For our task, we use the Infodemic Covid19 dataset \cite{infodemic} as one of the training dataset for our classifier. This dataset has 504 tweets in English and 218 tweets in Arabic, annotated with fine-grained labels related to disinformation about COVID-19. The labels answer seven different questions, which are related to negative effect and factual truthfulness of the tweet and we only consider the annotations for the first question, which is: \textit{"Does the tweet contain a verifiable factual claim?"}. For the interest of this task, we only utilize the English tweets of Infodemic Covid19 dataset.

\subsubsection{Indic Annotations}
In most of the existing tasks, detection of fake tweets have been predominantly done only in English. But in recent times, Twitter has seen a significant rise in regional language tweets\footnote{Source:  \url{https://tech.economictimes.indiatimes.com/news/internet/india-is-clocking-fastest-revenue-growth-for-twitter-india-md-manish-maheshwari/71999148}}. This formed our motivation to extend this task to Indic languages as well. For this task, the Indic languages of our choice are Hindi and Bengali. This choice is guided by the fact that these languages are the two most widely spoken languages in India\footnote{Source: \url{https://en.wikipedia.org/wiki/List\_of\_languages\_by\_number\_of\_native\_speakers\_in\_India}} and also the two most widely used Indic languages in the world. We now enlist the methods in which the Indic tweets were obtained and processed below.

We obtain the Bengali tweets from \cite{wdt0-ya78-20} which is a database of over 36,000 COVID related Bengali tweets, without any task-specific annotated labels. We randomly select 100 tweets from this database and annotate it with the same labeling schema followed by \cite{infodemic} for the first annotation question as mentioned above. We further augment the dataset with Bengali translations of the English tweets of the Infodemic dataset. We perform the translations using Google Translate API. Out of all the translations obtained we only keep those translations whose language is detected as Bengali by the same translation API.

For Hindi, we make use of the tweepy API \cite{tweepy} to scrape COVID related tweets from the Twitter interface. This was achieved by firing an API search with COVID related key-terms like "Covid", "Corona", etc. We follow the translation data augmentation process and the same labeling schema for Hindi tweets as well. Augmenting the dataset with machine translated texts adds noise to the dataset and helps in training a more robust model. The tweets in this dataset have been collected over a time period of March to May 2020 and records a balanced distribution of spam to non spam tweets which can be observed in the Tab. \ref{tab:dataset}.

\begin{table}[ht]
\centering
\begin{tabular}{|c|c|c|}
\hline
Language & Fake & Non-Fake \\ \hline
English  & 199  & 305      \\ \hline
Bengali  & 183  & 297      \\ \hline
Hindi    & 192  & 262      \\ \hline
Total    & 574  & 864      \\ \hline
\end{tabular}
\vspace{2mm}
\caption{Indic Misinformation Dataset Statistics}
\label{tab:dataset}
\end{table}

We also consider the features extracted from the Twitter user information, as a part of the dataset. For English, we make use of the user features provided by \cite{infodemic}. For Bengali and Hindi we have scraped the user information using the tweepy API and have provided it along the dataset. We have made our dataset and codes  publicly available to further the cause of research in this domain.\footnote{\url{https://github.com/DebanjanaKar/Covid19_FakeNews_Detection}}

\section{Proposed Approach} 

% name and Bert- explain 
% 1.25 page and picture
% training method
Our proposed approach is built on top of the method used in \cite{infodemic}. The main essence of the proposed approach lies in the features used for classification task and the different classifiers and their corresponding adaptation done for identifying the fake tweets. The details are described below.

\subsection{Extracted Features}
\begin{figure}[htb]
\centering
\includegraphics[scale=0.45]{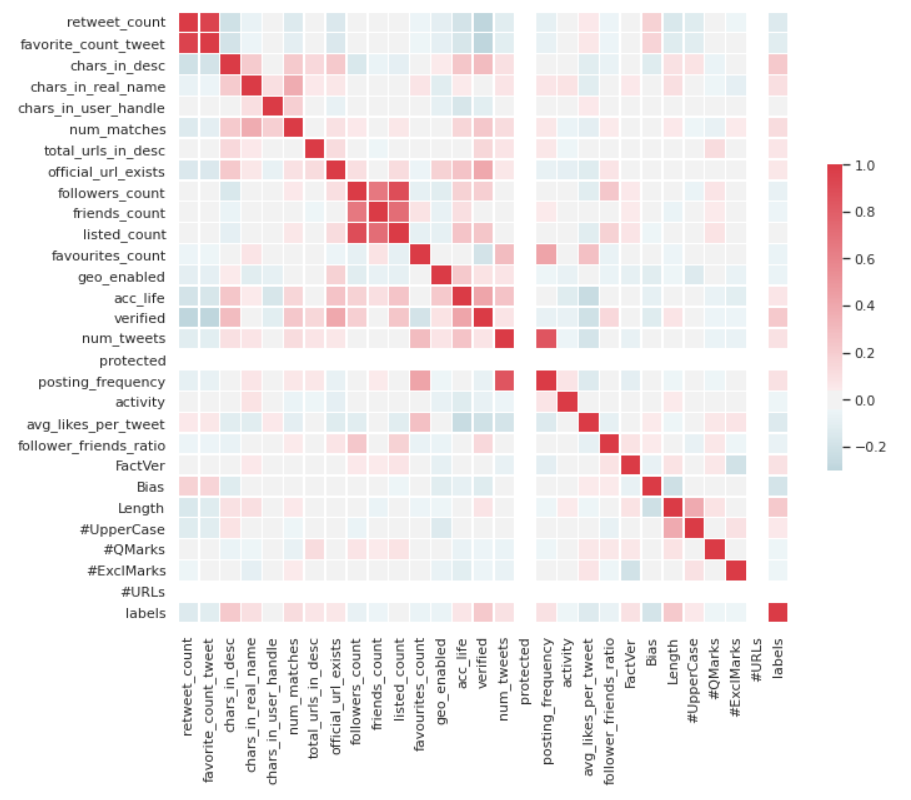}
\caption{Correlation of hand-crafted features with the class-labels show the effectiveness of the extracted features for the classification task.}
\label{fig:user_fet}
\end{figure}

We extract various textual and statistical information from the tweet text messages and user information separately, and analyse their role in the classification process. The different features are enlisted as follows :
\begin{enumerate}
  %\item \textbf{Multi-Lingual Bert Classifier Posterior Probability} (\textit{BERTPostProb}) - We fine-tune the multi-lingual BERT (uncased) classifier \cite{devlin2018bert} with our training set and detect if a tweet text is fake or not. We consider the class probability of the output label or the soft class label, as a features.\\
  %\item Keyphrase Embedding (\textit{keyphrembd}) - It has been reported in \cite{transferLr} that LASER embedding \cite{laser2018} provide extra richness to the feature set even when Bert embedding are being used. LASER embedding has been reported move effective than mBERT in calculating similarity between cross-lingual sentences. Following this fact, we extract the important keyphrase, by using MultiRake \cite{multirake}, from the tweet text and use the corresponding LASER embedding as a feature. We use a sub-model to downsize the number of LASER embedded features, so that it does not overshadow the important information from other feature groups. We use keyphrases, instead of the entire tweet, to give more emphasis on certain words (extracted keyphrase) which are more relevant to the context. 
  \item \textbf{Text Features} (\textit{tweettext}): We extract some of the twitter and textual features such as:
  
  \begin{enumerate}[label=\roman*]
  \item retweet\_count - The number of times a tweet has been retweeted
  \item favourite\_count - The number of likes received by a tweet.
  \item number\_of\_upper\_characters in a tweet
  \item number\_of\_question\_mark (s) in a tweet
  \item number\_of\_exclamation\_mark (s) in a tweet
  \end{enumerate}
  
  \item \textbf{Twitter User Features} (\textit{tweetuser}): "A man is known by the company he keeps." A considerable amount of literature in this task has already cited the immense importance of analysing the user's persona by extracting features from the user's Twitter profile. In our work, we extract the following 19 features from a user's profile, out of which 7 features are new additions with respect to the work in \cite{infodemic}: 
  \begin{enumerate}[label=\roman*]
  
  \item Chars{\_}in{\_}desc : The number of characters in user's description.  
  
  \item Chars{\_}in{\_}real{\_}name : The number of characters in user's real name. 
  
  \item Chars{\_}in{\_}user{\_}handle : The number of characters in user's handle on twitter.
  
  \item Num{\_}Matches (\textit{new}): Number of character matches in real name and username.
  
  \item Total{\_}URLs{\_}in{\_}desc (\textit{new}): Total number of URLs in user's description.
  
  \item Official{\_}URL{\_}exists (\textit{new}): Whether the user has an official URL or not.
  
  \item Followers{\_}count: Number of people that are following the user.
  
  \item Friends{\_}count: Number of people the user is following.
  
  \item Listed{\_}count: Number of lists to which the user has been added.
  
  \item Favourites{\_}count: Total number of likes the user has received throughout the account life.
  
  \item Geo{\_}enabled: Whether the user has allowed location access.
  
  \item Acc{\_}life (\textit{new}): Number of days since the account was created.
  
  \item Verified: Whether the user account is verified officially by twitter or not.
  
  \item Num{\_}tweet: Total number of tweets tweeted by the user.
  
  \item Protected: Whether the user account is protected or not.
  
  \item Posting{\_}frequency (\textit{new}): Number of tweets tweeted per day by the user.
  
  \item Activity (\textit{new}): Number of days since the user's latest tweet.
  
  \item Avg{\_}likes{\_}per{\_}tweet (\textit{new}): Total likes received by the user divided by total tweets tweeted by the user.
  
  \item Follower{\_}friends{\_}ratio: Followers' to friends' ratio.

\end{enumerate}

  \item \textbf{Fact verification score} (\textit{FactVer}) - We use the tweet text as a search query in Google search engine and consider the search results from reliable sources only. For this, we have manually selected 50 websites as reliable, which consists of popular news websites. We get the \textit{FactVer} score by taking the average of the Levenshtein distance between the tweet text and the titles obtained from the search results from these reliable links only.
  
  \item \textbf{Bias score} (\textit{Bias}) - An observed trait of fake tweets is that it often contains disturbing content which is malicious and in violation of Twitter's terms of use \cite{inuwa2018detection}. A 2018 Amnesty International report\footnote{Source:  \url{(https://www.amnesty.org.uk/press-releases/women-abused-twitter-every-30-seconds-new-study\#.XBjMprLxot0.twitter)}} shows women are abused on Twitter every 30 seconds, with racism, sexism and homophobia in the social media platform and usually such tweets contain untrue facts presented in a vindictive fashion. To explicitly make use of this knowledge, we use a Linear Support Vector Classifier (SVM) pre-trained on a very large annotated corpus. We use the pre-trained  profanity checker to obtain the probability of a tweet containing offensive language and use it as the 'bias score' in our model. As can be observed in the feature map above Fig. \ref{fig:user_fet}, the 'bias score' obtained shows a strong negative correlation with the fakeness of the tweets. 
 
  \item \textbf{Source Tweet Embedding} (\textit{TextEmbd}): The final and the most obvious feature in this task is the tweet itself. We consider the BERT/mBERT embeddings of the tweet text as the feature.
 \end{enumerate}

The correlation plots of the hand-crafted features with the class-label is shown in Fig. \ref{fig:user_fet}. It can be observed that not all features are helpful for the classification task. Nevertheless for now, we have considered all the features together, and keep automatic feature selection as a future scope of work.

\subsection{Classifier}

\begin{figure}[htb]
\centering
\includegraphics[scale=0.25]{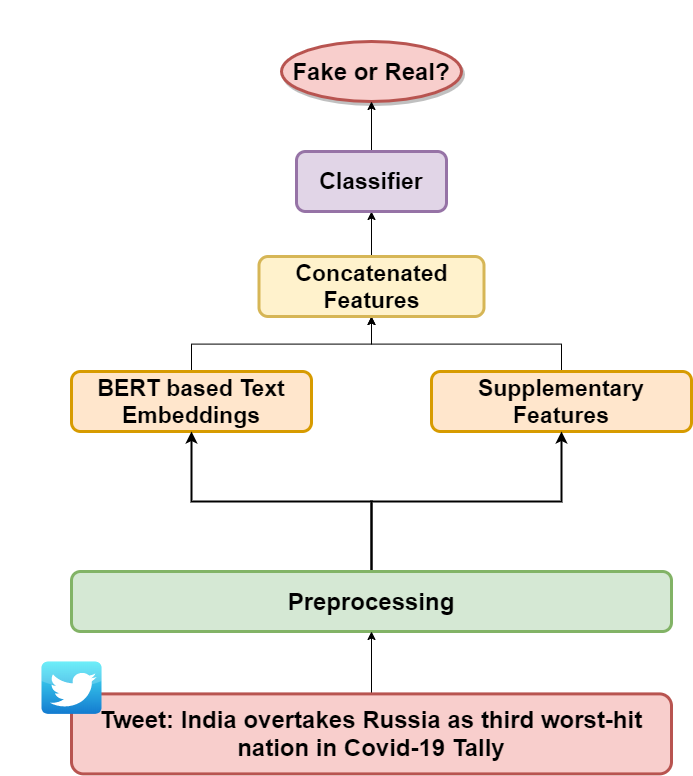}
\caption{Model (non Transformer) architecture used to detect fake tweets.}
\label{fig:architecture}
\end{figure}
Given a tweet, the above-mentioned features are extracted from it, various combinations of which are passed to the choice of classifier for the prediction task. The architecture of the proposed framework is shown in Fig. \ref{fig:architecture}. We use different feature combinations and classifiers for our purpose. We use Linear Support Vector Machine (SVM), Random Forest Classifier (RFC), Multi Layer Perceptron (MLP) and in-built multilingual BERT (mBERT, base cased) classifier (FC layer + softmax) for the classification task. We fine-tune the mBERT classifier by appending a layer (768 times 2) at the end followed by softmax and tune it for the classification task. The parameter details of the classifiers can be found in the experimental section. We have experimented with various combinations of classifiers and features and the noteworthy ones have reported in the next section.

%\begin{figure}[htb]
%\includegraphics[scale=0.45]{bert_archi.png}
%\caption{Architecture used for fine-tuning multilingual BERT model.}
%\label{fig:bert_ft}
%\end{figure}

\section{Experimental Details}
% 1/2 page
% Dataset (covid-infodemic) , Experiments
% human evaluation - (50- 100)
% Bert multilingual model 

We perform out experimentation on detecting fake tweets on our proposed \textit{Indic-covidemic fake tweet} dataset. The tweets are pre-processed, where the hashtags, URL, user-tags have been removed and the text is converted to lower case for uniformity. We first perform experimentation on English Tweets and extend it to suit the needs of the Indic tweets. The details are explained in the following sub-sections.

\subsection{English Tweet Classification - mono-Lingual classifier} 
We started our experimentation by classifying the English tweets to either Fake or Non-Fake class. We use the BERT model with pre-trained weights to get the sentence embedding of the tweet texts. We  used  80\%  of  the  dataset  for  training  and 20\%  for  testing. Pretrained BERT embeddings are being used with (or without) different feature combinations, to classify using Support Vector Machine (BERT\_SVM) with linear kernel (C=1, Gamma=1), Random Forest Classifier (BERT\_RFC) with 400 estimators, Multi-Layer Perceptron Neural Network (BERT\_MLP) classifier. MLP uses 30 neurons in the first hidden layer and 10 neurons in second hidden layer, with ReLU activation function and trained for 1000 epochs during training. We also use mBERT embedding with top classifier layer (mBERT\_NN) using FC layer (768x2) followed by softmax for classification purpose. Table \ref{tab:english_classifier} shows the performance analysis with various combinations of features and classifiers. We keep the notable feature combinations, while discarding the non-useful ones. The first row denotes the results obtained by Alam et. al. \cite{infodemic} using mBERT model, which is being considered for comparative study. 
%RFC gives the best result with the feature combination of tweet embedding and user features, which outperforms the current state-of-the-art \cite{infodemic} on Infodemic dataset. This is due to the fact that SVM suffers poorly with dataset which has much more number of features than that of the training set size. Also MLP requires high number of training samples during training, which may be missing in our settings.

\begin{table}[htb]
    \centering
    \begin{tabular}{|c|p{1in}|c|c|c|}
    \hline
    Model &  Features & \multicolumn{3}{c|}{Metric Scores}   \\ \cline{3-5}
     &  &      Prec. & Recall & F-score   \\ \hline \hline
     Infodemic \cite{infodemic} &  \textit{TextEmbd}  & - & - & 88.3 \\ \hline
     BERT\_SVM &   \textit{TextEmbd}  & 74 & 76 & 75 \\ 
     &  \textit{TextEmbd} + \textit{tweettext} & 50 & 31 & 38 \\  
     &  \textit{TextEmbd} + \textit{tweetuser} & 50 & 30 & 37.5 \\ \hline 
     BERT\_RFC &    \textit{TextEmbd}  & 73 & 75 & 74 \\ 
     &  \textit{TextEmbd} + \textit{tweettext} & 72 & 74 & 73 \\  
     &  \textit{TextEmbd} + \textit{tweetuser} & 88 & 89 & 89 \\ \hline 
     BERT\_MLP &  \textit{TextEmbd}  & 73 & 71 & 72 \\ 
     &  \textit{TextEmbd} + \textit{tweettext} & 64 & 62 & 63 \\  
     &  \textit{TextEmbd} + \textit{tweetuser} & 58 & 57 & 57.5 \\ \hline 
     mBERT\_NN &   \textit{TextEmbd}  & \textbf{87.17} & \textbf{91.89} & \textbf{89.47} \\ 
     (fine-tuned) &  \textit{TextEmbd} + \textit{tweettext} & 84.38 & 84.38 & 84.38  \\  
     &  \textit{TextEmbd} + \textit{tweetuser} & 90.32 & 87.5 & 88.88 \\ 
     &  \textit{TextEmbd} + \textit{FactVer} & 90.33 & 87.5 & 88.88 \\
     &  \textit{TextEmbd} + \textit{Bias} & 82.35 & 87.5 & 84.84 \\
     &  \textit{TextEmbd} + \textit{tweetuser} + \textit{FactVer} & 89.65 & 81.25 & 85.25 \\ \hline
    
    \end{tabular}
    \vspace{2mm}
    \caption{{\bf MONO-LINGUAL SETTING (ENGLISH TWEETS)}:{Precision, Recall and F-scores of our proposed method using different combinations of features and classifiers for English tweets.}}
    \label{tab:english_classifier}
\end{table}

\textit{Analysis} - We observe that adding of features always does not improve the performance of the classifier. This is not strange while dealing with hand-crafted features. Most of the user features are correlated to the class-label. Hence it is expected that addition of user features will boost the classifier performance. However, due to the shortcomings of SVM and MLP in dealing with small sample size problem, we rely on the metric numbers obtained by RFC (offers a piece-wise linear decision boundary in the feature space) and Fine-tuned in-build mBERT classifier (mBERT\_NN). We observe that \textit{TextEmbd}, \textit{tweetuser} and \textit{FactVer} are strong features and consider these for further analysis of Indic tweets. We emphasise on a high recall value as we want a low false positive value so that fake tweets gets filtered out to the best possible manner.

\subsection{Indic Tweet Classification - multi-Lingual classification}

\begin{table*}[h]
    \centering
    \begin{tabular}{|c|c|c|c|c|c|c|}
    \hline
     Evaluation type & \multicolumn{2}{|c|}{Language} & Features & \multicolumn{3}{c|}{Metric Scores}   \\ \cline{2-3}\cline{5-7}
      & Train & Test &  & Prec. & Recall & F-score   \\ \hline
     
       Cross-Domain & Eng+Hin+Ben & Hindi & \textit{TextEmbd} &  72.72 & 84.21 & 78.04 \\
       Data Augmentation & Eng+Hin+Ben & Hindi & \textit{TextEmbd}+\textit{FactVer} &  75.00  & 84.0  & 79.24 \\
       \cline{2-7} 
       &Eng+Hin+Ben & Bengali & \textit{TextEmbd} &  76.47 & 86.66 & 81.25 \\
       &Eng+Hin+Ben & Bengali & \textit{TextEmbd}+\textit{FactVer} & 73.5   & 83.33  & 78.12 \\
       \hline 
      Cross-Domain & Eng+Ben & Hindi & \textit{TextEmbd} &  70.30 & 95.80 & 81.09 \\    
       Zero-shot & Eng+Hin & Bengali  & \textit{TextEmbd} & 90.66 & 68.68 & 77.79 \\    
       & Hin+Ben & English  & \textit{TextEmbd} & 92.75 & 62.95 & 75.00 \\    \hline
     
    \end{tabular}
     \vspace{2mm}
    \caption{{\bf MULTILINGUAL SETTING (ENGLISH, HINDI, BENGALI):} Precision, Recall and F-scores of our proposed method using different combinations of features and classifiers (first three rows) and using different combinations of train and test sets for the best performing classifier for Indic languages (multi-lingual settings).}
    \label{tab:multilingual_classifier}
\end{table*}
We fine-tune mBERT and use text embedding (with and without \textit{FactVer}) as features. We use fine-tuned mBERT for the classification task (mBERT\_NN). Tab. \ref{tab:multilingual_classifier} shows the performance of different experiments in multi-lingual settings. We also test each of the language tweets separately in a zero-shot settings, where one of the language tweets are being used only as the test set. As mentioned by Alam et. al. \cite{infodemic}, transformers models are sensitive to the seed value specially when the training data is not very large. Following \cite{infodemic}. We experiment with 20 different random seed and fix one particular seed for the entire experimentation.

\textit{Analysis} - \textit{TextEmbd} is a strong feature which gives consistent performance in all the settings. This particular feature is tuned as per the FCLayer (in-build classifier of mBERT) which further favours it. Addition of \textit{FactVer} helps for Hindi, but not for Bengali tweets. However, we would like to conclude that it is a strong feature as well. It is to be noted, that the adaptation from English+Hindi to Bengali and English+Bengali to Hindi is encouraging. But when the classifier is tested on English tweets, while using Hindi+Bengali as train set, the performance deteriorates, This is due to the fact that the similarities between Hindi and Bengali is much more than Hindi-English or Bengali-English. 

\subsection{Ablation Study}
We study the cases where our proposed method acts dubiously, which are as follows:
\begin{itemize}
    \item Fake information from verified Twitter users - We have observed that the model gets confused in detecting false information when tweeted by a verified user. This can be overcome by considering similar examples during training, which will give less importance to the particular feature (verified) in \textit{tweetuser} feature set.
    \item Adapting model for English tweets in zero-shot learning setting - As mentioned above, the model does not generalize when English is the test set and the model has been trained on Hindi and Bengali tweets. This is reflected in the last row of Tab. \ref{tab:multilingual_classifier}. This is due to the immense linguistic difference of English with that of Hindi and Bengali. Since both Hindi and Bengali are Indo-Aryan languages, they are compatible with each other. This problem can be overcome by adding language specific layers in the architecture, which will take care of the linguistic features, specially in zero shot learning settings.
    \item Correlated features - Statistical classifiers do not always scale performance as the number of features increase, unlike deep neural networks. The hand-crafted features may seem intuitive at times, but may not always have the distinctive property to help in the classification task. We see this nature in our experiments as well. 
\end{itemize}
We would like to emphasis that these gaps are mostly due to less number of training samples and inherent drawbacks in the classifiers that we have adapted for the task. 

\subsection{API and Demo}
\begin{figure*}[h]
\centering
\includegraphics[width=16cm]{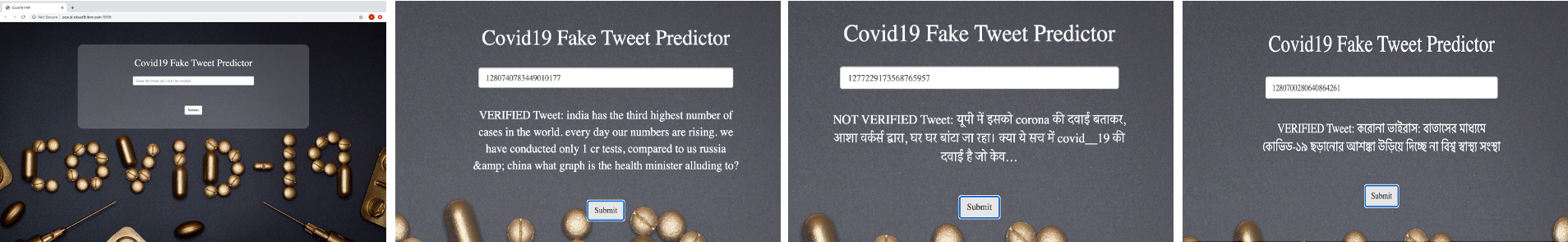}
\caption{Snapshots of the HTML page integrated with flask API which is used to predict tweet text using our proposed classifier model.}
\label{fig:api}
\end{figure*}
As a part of the project, we expose the model as a service through a flask API and static HTML page, where given the tweet id/text, it determines if the tweet is fake or not. Currently, this has been deployed in IBM intranet. It allows anybody to check if a given tweet is a fake one or not. Some of the API is shown in Fig. \ref{fig:api}.

%\section{Results and Analysis}
% 1 page 
% Metrics - Precision recall

\section{Concluding Remarks}
% 1/4 page
% future - using trusted sites to improve the model
% \begin{thebibliography}{00}
% This paper describes a novel method to detect fake tweets regarding COVID-19 in English, Hindi and Bengali languages. For this purpose, we have proposed a multilingual Indic tweet dataset, one of the first of its kind. We use multlingual BERT embedding along with various hand-crafted features for the classification task. 

In this work, we propose a multilingual approach to detect fake news about COVID-19 from Twitter posts for multiple Indic-Languages. 
In addition, we also created an annotated dataset, \textit{Indic-covidemic tweet dataset}, of Hindi and Bengali tweets for fake news detection. The proposed model is build on mBERT based embedding augmented with additional relevant features from Twitter to classify fake or genuine tweets. We show that our model reaches around 89\% F-score in fake news detection which supercedes SOTA scores for English dataset. We show that model trained with multiple Indic-Languages (our \textit{Indic-covidemic tweet dataset}) fake news dataset tweets shows improved performance which can be attributed to cross-lingual transfer learning as many Indic-Languages share similar syntactic constructs. Moreover, we establish first benchmark for two Indic languages, Hindi and Bengali. Our model achieves about 79\% F-Score in Hindi and 81\% F-score for Bengali Tweets. 
Further, with the focus to scale our model for other low resource Indic-Languages, we have proposed zero-shot learning approach for fake tweet detection. Our zero shot model achieves about 81\% accuracy in Hindi and 78\% accuracy for Bengali Tweets without any annotated data, which clearly indicates the efficacy of our approach. In future, we would like to add an automatic feature selection module which can take the most informative subset of handcrafted features for the classification task.

\bibliographystyle{IEEEtran}
\bibliography{covid}

\end{document}